\documentclass[letterpaper, 10 pt, conference]{ieeeconf}  

\IEEEoverridecommandlockouts       

\usepackage[backend=biber,
            url=false,
            isbn=false,
            doi=false,
            backref=false,
            style=ieee,
            natbib=true,
            mincitenames=1,
            maxcitenames=1,
            citestyle=numeric-comp,
            sorting=nyt,
            block=none]{biblatex}

\usepackage{algorithm}
\usepackage[noend]{algpseudocode}
\let\labelindent\relax
\usepackage[shortlabels]{enumitem}

\usepackage{graphics}
\usepackage{dsfont}
\usepackage[pdftex]{graphicx}
\usepackage{wrapfig}
\DeclareGraphicsExtensions{.pdf,.png,.jpg}
\usepackage{epsfig}
\usepackage[font={small}]{caption}
\usepackage{subcaption}
\usepackage[rightcaption]{sidecap}
\usepackage{pbox}

\usepackage{bigstrut}
\usepackage{mathtools}
\usepackage{amsmath, amssymb, amscd}
\usepackage{ wasysym } 
\usepackage{gensymb} 
\usepackage{nicefrac}       
\numberwithin{equation}{section} 

\DeclareMathAlphabet{\mathcal}{OMS}{lmsy}{m}{n}
\usepackage{textcomp} 

\usepackage{algorithm} 
\usepackage{algorithmicx, algpseudocode} 

\usepackage{array} 
\usepackage{tabularx}
\usepackage{multirow}
\usepackage{multicol}
\usepackage{booktabs}
\usepackage{tabulary}

\usepackage[utf8]{inputenc}
\usepackage[english]{babel} 
\usepackage{units}
\usepackage{bm}
\usepackage{bbm}
\usepackage{xspace}
\usepackage{flushend}
\usepackage{balance} 
\usepackage{csquotes}
\usepackage{makeidx}
\usepackage{blindtext}

\usepackage{ragged2e}
\usepackage{soul} 
\usepackage{subfiles} 

\usepackage{url}
\makeatletter
\g@addto@macro{\UrlBreaks}{\UrlOrds}
\makeatother
\usepackage{color}
\usepackage[usenames,dvipsnames,table,xcdraw]{xcolor}
\usepackage{pgfplots} 
\pgfplotsset{compat=newest}

\usepackage{marginnote}
\usepackage{soul} 

\usepackage[colorinlistoftodos]{todonotes}
\newcommand{\tocite}[1]{%
\textcolor{red}{[cite:\ifthenelse{\equal{#1}{}}{}{#1}?]}
}

\newcommand{\ignore}[1]{}

\renewcommand{\baselinestretch}{1.0}



\usepackage[pdfborder={0 0 0.5}]{hyperref}
\hypersetup{
    colorlinks=true,
    linkcolor=black,
    citecolor=black,
    filecolor=cyan,
    urlcolor=black
}

\def\one{\mathbbm{1}}

\newcommand{\vc}[1]{\bm{#1}}

\newcommand{\algname}{Iterative Reduction Of Non-planar Multiple cAble kNots (IRON-MAN)}

\newcommand{\algabbr}{IRON-MAN}

\title{\LARGE \bf
Disentangling Dense Multi-Cable Knots
}

\author{Vainavi Viswanath$^{*1}$, Jennifer Grannen$^{*1}$, Priya Sundaresan$^{*1}$, Brijen Thananjeyan$^{1}$, Ashwin Balakrishna$^{1}$,\\ Ellen Novoseller$^{1}$, Jeffrey Ichnowski$^{1}$, Michael Laskey$^{2}$, Joseph E. Gonzalez$^{1}$, Ken Goldberg$^{1}$ 
\thanks{$^{1}$AUTOLAB at the University of California, Berkeley}
\thanks{$^{2}$Toyota Research Institute}
\thanks{*equal contribution}
}

\begin{document}

\maketitle

\begin{abstract}
Disentangling two or more cables requires many steps to remove crossings between and within cables. We formalize the problem of disentangling multiple cables and present an algorithm, \algname{}, that outputs robot actions to remove crossings from multi-cable knotted structures. We instantiate this algorithm with a learned perception system, inspired by prior work in single-cable untying that given an image input, can disentangle two-cable twists, three-cable braids, and knots of two or three cables, such as overhand, square, carrick bend, sheet bend, crown, and fisherman's knots. \algabbr{} keeps track of task-relevant keypoints corresponding to target cable endpoints and crossings and iteratively disentangles the cables by identifying and undoing crossings that are critical to knot structure. Using a da Vinci surgical robot, we experimentally evaluate the effectiveness of \algabbr{} on untangling multi-cable knots of types that appear in the training data, as well as generalizing to novel classes of multi-cable knots. Results suggest that \algabbr{} is effective in disentangling knots involving up to three cables with 80.5\% success and generalizing to knot types that are not present during training, with cables of both distinct or identical colors.



\end{abstract}

\section{Introduction}
Knots composed of multiple ropes and cables are found in many environments, including electronic cords in homes, offices, and concert stages, electrical wiring in warehouses, ropes in sailboats and ships, and cables in manufacturing settings~\cite{mayer2008system, van2010superhuman, yamakawa2007one, sanchez2018robotic}. Furthermore, rope and cable disentangling can be critical in life-saving systems for search-and-rescue operations and disaster response~\cite{merlet2010portable, kempf2011fiber}, where disentangling knots is crucial for task success. Designing robust systems for disentangling 1D deformable objects, which we refer to as ``cables,'' is challenging. In this paper, we propose methods that leverage the graphical structure of multi-cable knots to address this challenge.

This work considers cable \emph{disentangling}, that is, separating two or more cables that are knotted or twisted together. While prior work has studied the problem of untying knots in a single cable~\cite{lui2013tangled, grannen2020untangling}, disentangling multiple cables involves new challenges. First, multi-cable knots can contain more crossings than single-cable knots,  especially in cases where the cable is stiff and has a large turning radius: multiple cables can intertwine with each other with minimal turning, while single cables require a smaller turning radius to achieve the same number of intersections. Second, perception is complicated by the presence of additional cable endpoints, higher-density configurations with little space between adjacent cable segments, and a greater potential for occlusion. Third, the increased number of cable endpoints in multi-cable systems makes it difficult to track untangling progress along specific cables. Fourth, the mechanics of cable manipulation are also more complicated with multiple cables, as crowding of cables can impede reachability. 

\begin{figure}[t!]
\centering
\includegraphics[width=1.0\linewidth]{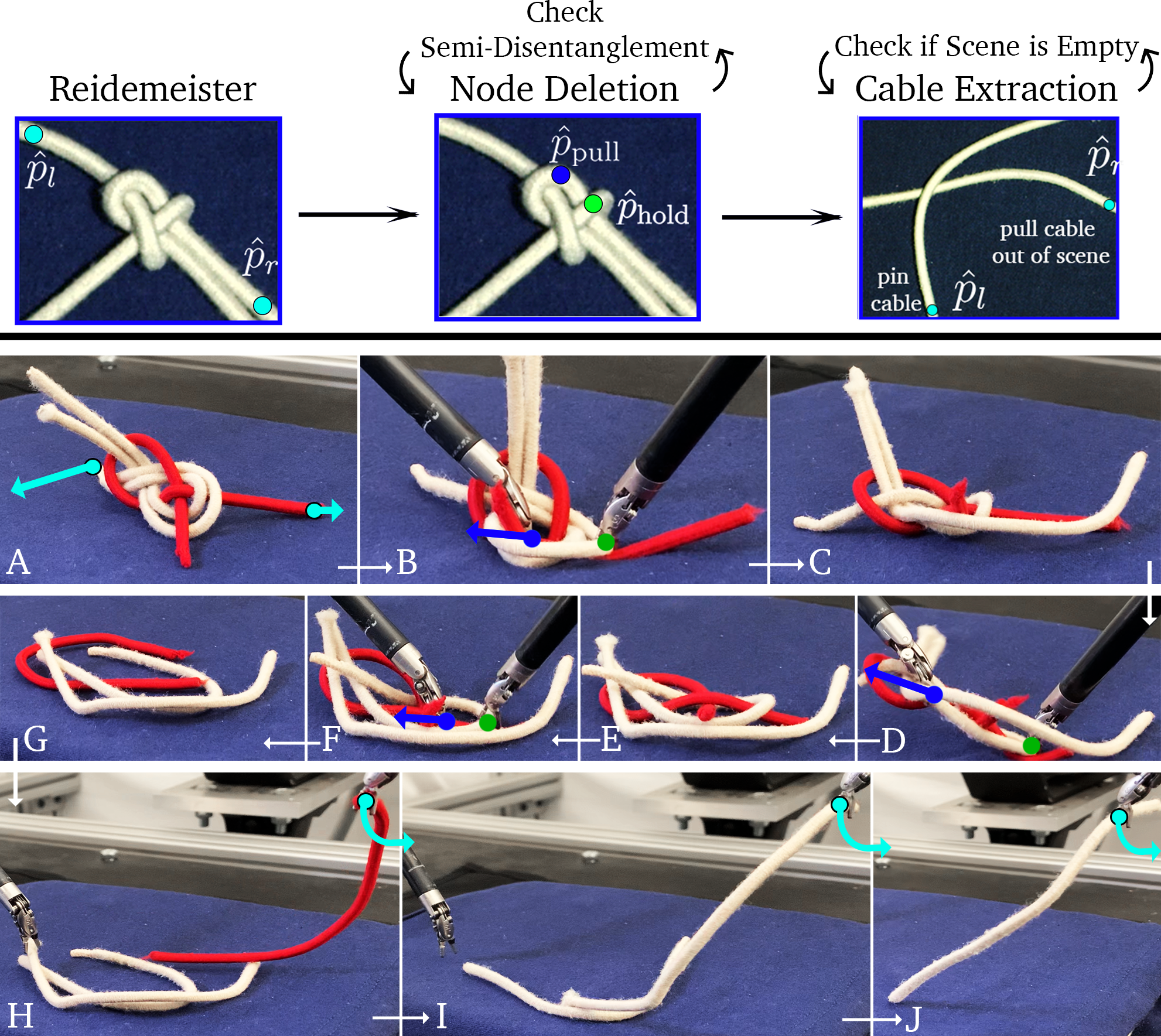}
\caption{\textbf{Overview of \algabbr{}}: \algabbr{} (Iterative Reduction Of Non-planar Multiple cAble kNots) is an algorithm for disentanglement of several knotted cables. We present a sequence of moves planned by \algabbr{} on a three-cable Carrick Bend knot. Following an initial Reidemeister (straightening) move (A) which pulls opposing cable endpoints apart, \algabbr{} takes several Node Deletion (loosening) moves (B-C, D-E, F-G) to reduce inter and intra-cable crossings. Finally, we take three Cable Extraction (removal) moves (H-J) to isolate and remove each cable. 
}
\end{figure}

Prior methods for single-cable untangling~\cite{grannen2020untangling} cannot be straightforwardly adapted to the multi-cable setting, which involves cable configurations of greater complexity and density and presents new challenges for perception and action selection. \citet{grannen2020untangling} present a geometric algorithm for untangling a single cable that iteratively undoes each cable crossing, starting from one cable endpoint and working toward the other. A naive approach to multi-cable untangling could consider each cable sequentially to untie knots within each individual cable. However, multi-cable systems have many endpoints, which cannot always be easily mapped to specific cables. Furthermore, the presence of inter-cable crossings in multi-cable knots further complicates planning of disentangling actions compared to the single-cable setting.

We propose \algname{}, an algorithm for disentangling multiple cables given a graphical representation of the knot structure. \algabbr{} distills full configuration information by defining a disentangling hierarchy over cable crossings to generate bilateral disentangling actions. \algabbr{} then prioritizes disentangling crossings integral to the knot structure. To implement \algabbr{} from visual observations, we employ keypoint regression methods~\cite{grannen2020untangling} to learn pull and hold keypoints. We plan manipulation actions over the learned keypoints with five manipulation primitives: two from knot theory and prior work~\cite{lui2013tangled, grannen2020untangling}, two from \citet{nonplanar-cable-untangling}, and a novel move introduced in this work.

This paper contributes: (1) a formulation of the multi-cable disentangling problem; (2) \algname{}, a novel geometric algorithm for disentangling multiple cables, (3) an instantiation of \algabbr{} from image input that extends the perception-driven single-cable planner from \citet{grannen2020untangling}; and (4) physical cable disentangling experiments on multi-cable knots of three difficulty tiers consisting of combinations of different types of multi-cable knots, twists, and braids. Experiments suggest that the physical implementation of \algabbr{} can completely disentangle all cables in scenes containing up to three cables with 80.5\% success.


\section{Background and Related Work}
\label{sec:related-work}
Deformable object manipulation has gained significant traction in the robotics research community in recent years~\cite{herguedas2019survey, sanchez2018robotic, billard2019trends}. Yet, deformable manipulation remains challenging due to complex dynamics, visually uniform appearances, and self-occlusions. Prior work has focused on developing long-horizon perception-driven planners for deformable manipulation, but tend to focus on handling a single object instance such as a cable, bag, or piece of fabric. In contrast, this work embeds awareness of both within and cross-instance geometry into a planner for multi-cable disentanglement.

\subsection{Deformable Object Manipulation}
Much existing work on robot manipulation of nonrigid objects infers a partial or full state estimation of a deformable object. In particular, \citet{yan2020self} and \citet{lui2013tangled} infer rope representations as sparse point sets from learned and analytical methods, respectively. Similarly, \citet{chi2019occlusion} demonstrate cloth and rope tracking based on template shape registration. To generalize single-instance perception models to multi-instance scenes, prior approaches rely on instance segmentation, which is difficult to achieve in the setting of homogeneous, tangled cables. \citet{florence2018dense} propose mapping object images to a dense pixel-wise descriptor embedding with which to recover object pose in both single and cluttered multi-instance scenes for semantic grasping. Dense descriptors have proven effective in rope knot-tying ~\cite{sundaresan2020learning} and cloth folding and smoothing ~\cite{ganapathi2020learning}, but global descriptors lack robustness to severe deformation and occlusion, as encountered with multiple overlapping cables.

In contrast, other approaches circumvent state estimation by performing end-to-end visuomotor learning for goal-conditioned tasks. These include rope shape-matching and knot-tying  by learning dynamics models \cite{nair2017combining, pathak2018zero}; cable vaulting by behavioral cloning ~\cite{zhang2020robots}; cloth smoothing and folding from video prediction models ~\cite{ebert2018visual,hoquevisuospatial}, latent dynamics models ~\cite{lin2020softgym}, reinforcement learning ~\cite{lee2020learning}, and imitation learning~\cite{seita2018deep, seita2019deep}; and bag manipulation by inferring spatial displacements~\cite{seita2020learning}. While general, these algorithms do not leverage the geometric structure specific to the cable manipulation problem, which makes them difficult to apply to highly complex tasks such as cable disentangling, in which fine-grained perception and manipulation is critical for success.

\subsection{Cable Untangling Methods}
Prior work has studied the task of single-cable untangling from both loose~\cite{lui2013tangled} and dense~\cite{grannen2020untangling, nonplanar-cable-untangling} initial configurations, where dense configurations lack space between crossings. \citet{lui2013tangled} propose modeling a cable configuration via a graphical abstraction representing cable crossings and endpoints, and approximate this model from RGB-D input through analytical feature engineering. This method assumes reliable segmentation of crossings to construct the graph, and as a result does not readily adapt to untangling dense, non-planar knots within or across cables. \citet{grannen2020untangling} define a single-cable untangling algorithm, HULK (Hierarchical Untangling from Learned Keypoints), which rather than explicitly reconstructing a cable graph as in \cite{lui2013tangled}, learns to predict untangling actions from RGB image observations. While HULK addresses semi-planar cable knot untangling, it does not account for disentangling, or separation of multiple cables in non-planar knots. HULK also does not apply out-of-the-box to the multi-cable setting due to ambiguity in selecting knot-loosening actions when there are many cable endpoints from which to trace under-crossings. Lastly, \citet{nonplanar-cable-untangling} define single-cable algorithms LOKI (Local Oriented Knot Inspection) and SPiDERMan (Sensing Progress in Dense Entanglements for Recovery Manipulation), which plan grasp refinement steps such as recentering the knot in the workspace and removing the knot from the gripper jaws when it becomes wedged there. Similarly to HULK, LOKI and SPiDERMan must be modified for the multi-cable scenario to account for crossings containing $>$3 cable segments and crossings involving cables of similar or different colors.
In this work, we extend the graphical abstraction from~\cite{lui2013tangled} to propose \algabbr{}, a scheme for systematically disentangling multiple cables that extends HULK, LOKI, and SPiDERMan to accommodate inter-cable crossings, resolve ambiguity in action selection and termination, and reason about the many cable endpoints.

\section{Task Formulation}

\begin{figure*}[!htbp]
\centering
\includegraphics[width=0.85\linewidth]{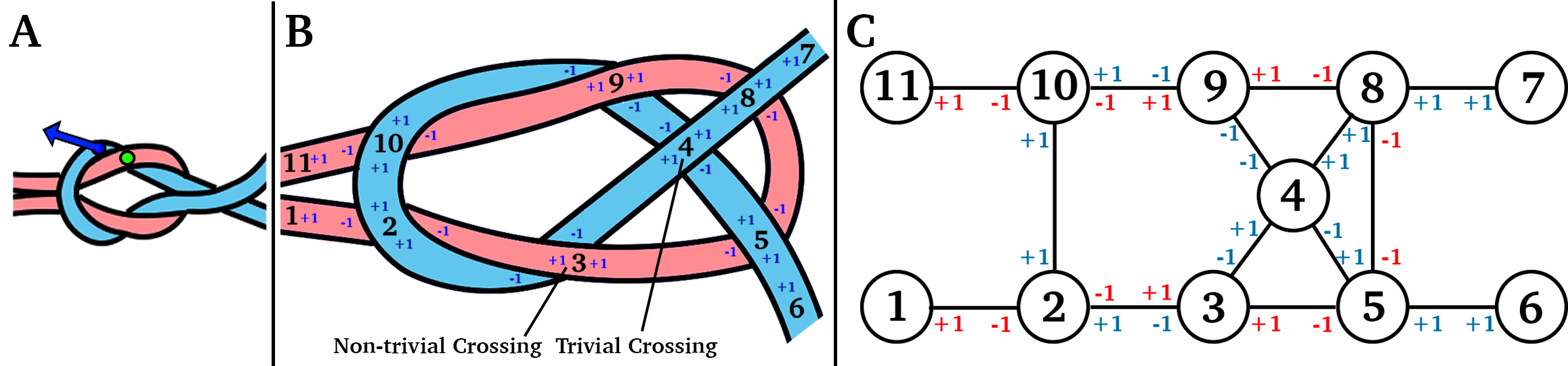}
\caption{\textbf{Graph Representation}: Provided a dense initial square knot (A), we take a Node Deletion move specified by hold (green) and pull (dark blue) keypoints, yielding a looser configuration shown in (B). We use a graphical abstraction to model the state of intertwined cables, extending previous work on modelling single-cable configurations ~\cite{lui2013tangled, grannen2020untangling, nonplanar-cable-untangling}. In this graph, endpoints and intra/inter-cable crossings constitute nodes, and edges denote over (+) and under (-) crossings, shown in (C). We prioritize removing crossings that are \emph{non-trivial}, such as (3), rather than \emph{trivial} ones such as (4), which can be easily undone by a Reidemeister move without significantly loosening the configuration. 
}
\label{fig:graph}
\vspace{-0.3cm}
\end{figure*}

\label{sec:problem-statement}
A bilateral robot aims to disentangle a knotted configuration of $n$ cables, where $n > 1$ initially, by removing one crossing at a time. The objective is to reach a fully disentangled state with no crossings, as defined below. In this section, we define the workspace and notation for cable untangling (Section~\ref{subsec:workspace}), present a graphical abstraction to represent the state of the cables in the workspace (Section~\ref{subsec:config-graph}), discuss the set of actions which can be used to manipulate the cables (Section~\ref{subsec:actions}), and formalize the objective of the multi-cable disentangling problem (Section~\ref{subsec:PS}).

\subsection{Workspace Definition}
\label{subsec:workspace}
We define a Cartesian $(x,y,z)$ coordinate frame for the workspace and assume it contains a bilateral robot. Without loss of generality, we assume that the manipulation surface lies in the $xy$-plane. For planning purposes, we define three points, $\vc{w}_l$, $\vc{w}_r$, $\vc{w}_c \in \mathbb{R}^3$, respectively located on the left bound, right bound, and center of the manipulation surface. This workspace setup resembles that in \citet{grannen2020untangling}, but differs from prior work in that the scene now contains $n$ cables that can be knotted or twisted together rather than 1. With these $n$ cables occupying the same workspace as a single cable in prior work, we now require improved cable slack management in the workspace. 
We assume the existence of a termination area to the right of the manipulation workspace, into which cables are relocated as they become fully disentangled. The termination area is centered at the point $\vc{p}_{\text{term}} \in \mathbb{R}^3$ and can be reached by the robot's grippers.

We define \emph{intra-cable crossings} to be crossings that only involve a single cable,
while \emph{inter-cable crossings} are crossings involving at least two cables.
The structure of crossings between cables is not directly observable, and must instead be inferred from RGB images. At time $t$, an image $I_t$ is inputted to the algorithm to generate manipulation actions.

\subsection{Configuration Graph}\label{subsec:config-graph}
Previous works define a graph structure to concisely represent the configuration space of single-cable knots~\cite{grannen2020untangling, lui2013tangled,nonplanar-cable-untangling} and introduce algorithms that operate on this compressed representation to perform untangling. This work extends these models to represent $n$-cable knots. We define a graph containing vertices (also referred to as nodes) $\mathnormal{v} \in \mathnormal{V}$, which represent any cable endpoints and crossings in the structure, and edges $\mathnormal{e} \in \mathnormal{E}$, which represent cable segments (with no crossings) that connect two vertices, and are denoted as $\mathnormal{e} = (\mathnormal{u},\mathnormal{v})$ for $\mathnormal{u},\mathnormal{v} \in \mathnormal{V}$.
While the graph representations in prior work limit crossings to 2 or 3 cable segments \cite{grannen2020untangling, nonplanar-cable-untangling}, in this work, a crossing (vertex $v$) can have $k$ segments for any $k \ge 1$, such that $2k$ cable segments extend from the crossing, and the corresponding vertex $\mathnormal{v}$ has a degree of $2k$. The graph vertices do not distinguish between intra-cable and inter-cable crossings. Endpoint vertices differ in that they always have a degree of one. Additionally, we annotate every adjacent (vertex, edge) pair with a label $X(v,e) \in \{-1,\ldots,-(k-1)\}\cup\{+1\}$, where $k$ is the number of segments in the crossing at vertex $v$, according the definition below:
\begin{equation}
\label{eq:annotations}
X(v, e) = 
\begin{cases} 
+1 & \parbox{16em}{if  $v$ is an endpoint or if $e$ crosses over all other edges at $v$}  \\
-m & \text{if $e$ crosses under $m$ edges at $v$}
\end{cases}
\end{equation}
Intuitively, $X(v, e)$ indicates the depth of edge $e$ at the crossing represented by vertex $v$. Observe that this graph can have multiple edges, as illustrated in Figure~\ref{fig:graph}, but no two pairs of contiguous edges will have the same two annotations.

\subsection{Action Space}
\label{subsec:actions}
We define the action space at time $t$ in terms of global workspace coordinates:
\[ \vc{a}_{t,\text{right}} = (x_{t,\text{r}}, y_{t,\text{r}}, \theta_{t,r}, \Delta x_{t,\text{r}}, \Delta y_{t,\text{r}}, \one_\mathrm{grasp})\] 
\[ \vc{a}_{t,\text{left}} = (x_{t,\text{l}}, y_{t,\text{l}}, \theta_{t,l}, \Delta x_{t,\text{l}}, \Delta y_{t,\text{l}}, \one_\mathrm{grasp}).\]

Gripper $k \in \{\text{right},\text{left}\}$ moves to coordinates $(x_{t,k}, y_{t, k})$ at a top-down grasp orientation $\theta_{t,k}$ about the $z$-axis and with a $30^{\circ}$ approach angle relative to the z-axis. Upon grasping, the jaw moves by $(\Delta x_{t, k}, \Delta y_{t, k})$ and releases its hold. When $\one_\mathrm{grasp} = 1$, these motions are performed with the gripper jaws closing onto the cable, while $\one_\mathrm{grasp} = 0$ indicates that the arms execute the same motions with the gripper jaws remaining open throughout, preventing a secure grasp. 

The actions $\vc{a}_{t,\text{right}}$ and $\vc{a}_{t,\text{left}}$ are executed simultaneously by both arms, and single-arm actions can be performed by setting the other arm's action to null. We will define specialized motion primitives in terms of this general action definition in Sections \ref{sec:motion primitives} and \ref{sec:hulk untangling}.

We assume access to a transformation between pixel coordinates $(p_x, p_y)$ and global positions $(x, y, z)$. Because the perception systems operate in pixel space, we will describe positions in terms of pixels, overloading the action notation to use pixel coordinates as well as workspace positions.

\subsection{The Multiple Cable Disentangling Problem}
\label{subsec:PS}
The objective of the multi-cable disentangling problem is to remove all intra-cable and inter-cable crossings in the scene with a minimal number. In terms of the knotted structure's graphical representation, the goal is to reach a configuration graph with two vertices per cable, one corresponding to each endpoint, such that the two vertices belonging to each cable are connected by an edge with positive annotations on both ends. At time $t$, the algorithm receives an image observation $I_t$ and outputs a linear, bilateral action $\vc{a}_t = (\vc{a}_{t,\text{right}}, \vc{a}_{t,\text{left}})$.

\section{Preliminaries}
\label{sec:Algorithm}
\subsection{Assumptions}
We make the following assumptions: 1) \textit{cables distinguishable}: the cables are visually distinguishable from the background via color thresholding, but need not be distinguishable from each other;
2) \textit{visible endpoints}: at least two endpoints are visible in the initial cable configuration; 3) \textit{linear pull actions sufficient}: we assume all cables are within reachable limits of the robot, so the robot can successfully perform grasping and pulling actions. Unlike \citet{grannen2020untangling}, who assume a semi-planar knot structure---i.e., at most two cable segments per crossing---we allow non-planar knots, where more than two cable segments can be involved in each intersection.

\subsection{Physical Disentangling System}
\label{sec:physical dis}
In this section, we describe three methods developed in prior work for single-cable untangling: HULK \cite{grannen2020untangling}, LOKI, and SPiDERMan \cite{nonplanar-cable-untangling}. 
\algabbr{} builds upon these methods to define an algorithm for untangling multiple cables. The planned grasps are executed using the grasp refinement steps from LOKI.

\subsubsection{HULK---Hierarchical Untangling from Learned Keypoints}
HULK~\cite{grannen2020untangling} predicts the locations of four task-relevant keypoints as pixel coordinates in the scene to plan motion primitives. For a single cable, HULK senses:
\begin{enumerate}
    \item $\vc{\hat{p}}_l$: the estimated pixel coordinate of the left endpoint.
    \item $\vc{\hat{p}}_r$: the estimated pixel coordinate of the right endpoint.
    \item $\vc{\hat{p}}_{\text{hold}}$: the estimated pixel coordinate of the topmost segment of cable at the first undercrossing $c$ from the rightmost endpoint.
    \item $\vc{\hat{p}}_{\text{pull}}$: the estimated pixel coordinate of a cable segment edge labeled $-i$ ($i > 1$) exiting the first undercrossing $c$ traced from the rightmost endpoint.
\end{enumerate}
For each keypoint $\vc{\hat{p}}$, HULK learns a mapping $f: \mathbb{R}^{640 \times 480 \times 3} \mapsto \mathbb{R}^{640 \times 480 \times 1}$ which maps an RGB image to a heatmap centered at $\vc{\hat{p}}$. HULK is trained from images, where given a hand-specified keypoint annotation for an image, we compute a ground truth 2D Gaussian distribution centered at the annotated pixel and with $\sigma=8 \text{px}$, similarly to~\cite{nonplanar-cable-untangling, grannen2020untangling}. 
In Section~\ref{sec:disentangling actions}, we redefine these keypoints analogously for the multi-cable setting.


\subsubsection{LOKI---Local Oriented Knot Inspection}
\label{sec:loki}
LOKI~\cite{nonplanar-cable-untangling} is a low-level grasp planner that computes a robust grasp by refining an inputted coarse grasp location to center it on a cable and infer a grasp orientation that is orthogonal to the cable's path. These refinements prevent near-miss grasps. LOKI maps a local crop of an image in $\mathbb{R}^{200 \times 200}$ centered at one of the keypoints to 1) $\theta$: an angle about the $z$-axis for top-down grasp orientation, and 2) $(p_{\text{off},x}, p_{\text{off},y})$: a local offset in pixel space to recenter the keypoint along the cable width. For an input keypoint $\vc{\hat{p}}$, we let $\vc{\tilde{p}}$ denote its refined grasp location (in pixels) and $\tilde{\theta}$ (from $0^{\circ}$ to $180^{\circ}$) denote its grasp orientation.

\subsubsection{SPiDERMan---Sensing Progress in Dense Entanglements for Recovery Manipulation}
SPiDERMan \cite{nonplanar-cable-untangling} addresses four manipulation failure modes observed in ~\citet{grannen2020untangling}. Here we implement SPiDERMan's recovery manipulation primitives \textbf{Wedge Recovery} and \textbf{Re-posing (translation)}, which are relevant to the task of disentangling multiple cables.
\textbf{Wedge Recovery} detects when a gripper is wedged between cable segments. If gripper $k$ is wedged between cable segments, it moves to the center of the workspace $w_c$ and opens its jaws. The other gripper pins the cable while the arm corresponding to gripper $k$ returns to its home position. When SPiDERMan detects that the cable mass is near workspace limits, it executes a \textbf{Re-posing move (translation)} to grasp the cable at its center and returns it to the center of the workspace $w_c$.

\subsection{Motion Primitives}
\label{sec:motion primitives}
We extend two motion primitives from \citet{grannen2020untangling} to fully disentangle cables: Reidemeister moves and Node Deletion moves. We use LOKI to center each grasp along the cable width with an improved grasping orientation.

\subsubsection{Reidemeister moves}
\textbf{Reidemeister moves} grasp the left endpoint at $\vc{\tilde{p}}_{l}$ with orientation $\tilde{\theta}_l$ and pull the cable to a predefined location $\vc{w}_l$ at the left side of the workspace. The right gripper similarly grasps the right endpoint at $\vc{\tilde{p}}_r$ with orientation $\tilde{\theta}_r$ and pulls the cable to a predefined location $\vc{w}_r$ at the right side of the workspace. This move eliminates trivial crossings not involving a knot and disambiguates the cable's configuration cable by spreading it apart.
The actions are described as follows:
\begin{flalign*}
\vc{a}_{t,l} &= (\tilde{p}_{l,x}, \tilde{p}_{l,y}, \tilde{\theta}_{l}, w_{x,l}-\tilde{p}_{l,x}, w_{y,l}-\tilde{p}_{l, y},1) \\
\vc{a}_{t,r} &= (\tilde{p}_{r, x}, \tilde{p}_{r, y}, \tilde{\theta}_{r}, w_{x,r}-\tilde{p}_{r, x}, w_{y,r}-\tilde{p}_{r, y},1).
\end{flalign*}

\subsubsection{Node Deletion moves}
After HULK identifies the first undercrossing traced from the rightmost endpoint and LOKI refines the grasps, a Node Deletion move attempts to pull out part of a cable segment underneath the topmost segment to eliminate undercrossings.
In a \textbf{Node Deletion move}, the right gripper holds a cable segment at $\vc{\tilde{p}}_{\text{hold}}$ while the left gripper grasps a cable segment at $\vc{\tilde{p}}_{\text{pull}}$ and pulls out the cable slack underneath $\vc{\tilde{p}}_{\text{hold}}$. LOKI predicts each grasp rotation $\tilde{\theta}_{\text{hold}}$ and $\tilde{\theta}_{\text{pull}}$, to obtain actions described as: 
\begin{flalign*}
\vc{a}_{t,\text{hold}} &= (\tilde{p}_{\text{hold},x}, \tilde{p}_{\text{hold},y}, \tilde{\theta}_{\text{hold}},0, 0, 1) \\ 
\vc{a}_{t, \text{pull}} &= (\tilde{p}_{\text{pull},x}, \tilde{p}_{\text{pull},y},\\ &\hspace{0.2in} \tilde{\theta}_{\text{pull}}, \tilde{p}_{x,\text{pull}}-\tilde{p}_{\text{hold},x}, \tilde{p}_{\text{pull},y}-\tilde{p}_{\text{hold},y}, 1).
\end{flalign*}
\section{Methods}\label{sec:methods}
To the best of our knowledge, prior work has only considered single-cable untangling, and the resulting algorithms cannot be directly employed for disentangling multiple cables. We present a novel algorithm, \algabbr{}, for disentangling multiple cables using the graph representation of the knot configuration defined in Section~\ref{subsec:config-graph}. We then discuss perception methods and a novel manipulation primitive for managing excess cable slack to instantiate \algabbr{} to physically untangle multiple cables.

\subsection{Graph-Based Disentangling Action Planning}
\label{sec:disentangling actions}
\begin{figure}
\centering
\includegraphics[width=1.0\linewidth]{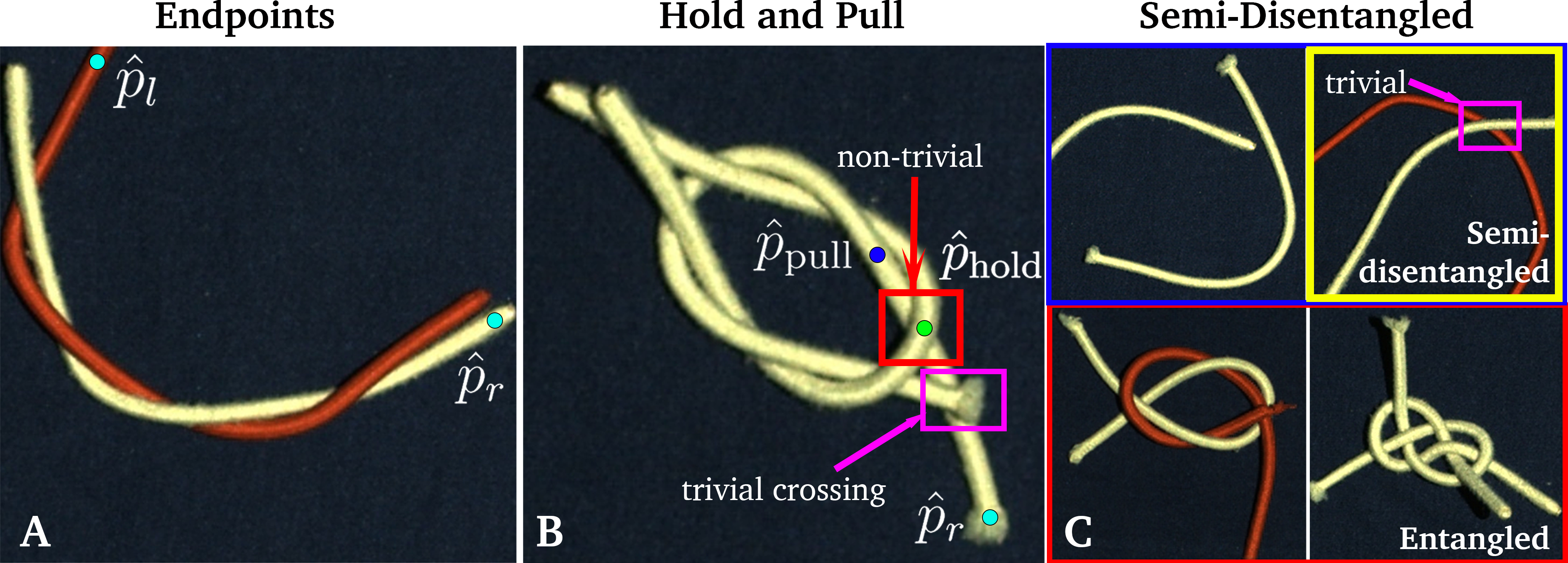}
\caption{\textbf{Physical Implementation of \algabbr{}}: Image A depicts the endpoints used for Reidemeister moves: $\vc{\hat{p}}_r$ and $\vc{\hat{p}}_l$ on cables $c_1$ and $c_2$, respectively. The left endpoint annotation corresponds not to the leftmost endpoint in the scene (on cable $c_1$), but rather to the leftmost endpoint of cable $c_2$ because $\vc{\hat{p}}_r$ is on cable $c_1$. Image B depicts the hold and pull keypoints, $\vc{\hat{p}}_{\text{hold}}$ and $\vc{\hat{p}}_{\text{pull}}$, relative to the first non-trivial crossing from the right endpoint. The first crossing from the rightmost endpoint is a trivial crossing, and is skipped when traversing from the right endpoint to annotate a Node Deletion move. Image C presents semi-disentangled and entangled configurations. Although the image in the yellow highlighted box still contains a crossing, it is trivial and thus acceptable by our definition of semi-disentanglement.}
\label{fig: labelling}
\end{figure}

We present \algabbr{}, an algorithm for disentangling multiple cables knotted or twisted together. \algabbr{} assumes access to only the knot's implicit graph structure and disentangles $n$ cables by removing crossings repeatedly. Multi-cable disentangling requires reasoning about the $2n$ endpoints, increased number of crossings, and complex cable slack management. 

\subsubsection{Multi-cable Reidemeister Moves}
With $n > 1$ cables, the scene contains $n$ right endpoints and $n$ left endpoints. \algabbr{} first locates the rightmost endpoint node $v_r$, which belongs to some cable $c_i$, $i \in \{1, ..., n\}$. After defining $v_r$, the leftmost endpoint node $v_l$ is defined as the leftmost endpoint belonging to some cable $c_j$, where $j\neq i$ and $j \in \{1, ..., n\}$ (Fig. \ref{fig: labelling}). When only a single cable $c_k$ remains in the scene so that $n = 1$, we define $v_r$ and $v_l$ to be the right and left endpoints of $c_k$, breaking ties arbitrarily. Reidemeister moves are performed on the endpoints $v_r$ and $v_l$. 

\subsubsection{Multi-Cable Node Deletion Moves}
Due to the increased cable length in the scene, each crossing removal requires complicated slack management, as we must successfully perform multiple Node Deletion actions without introducing new crossings when pulling cable slack through existing crossings. \algabbr{} categorizes crossings as either \emph{non-trivial} or \emph{trivial} when determining which crossings to remove. \emph{Non-trivial crossings} are integral to maintaining the knot structure, while \emph{trivial crossings} are not and can be undone by performing a Reidemeister move. Removing a trivial crossing from a configuration does not change the number and types of knots present in the configuration, and as a result does not reduce the overall configuration density. \algabbr{} manages physical cable slack effectively and efficiently by only undoing non-trivial crossings, while Reidemeister moves remove the trivial crossings. \algabbr{} traverses the graph from the rightmost endpoint $v_r$ and performs a Node Deletion move on the first non-trivial undercrossing it encounters. 

\subsubsection{Algorithm Summary}
\algabbr{} disentangles $n$-cable knots given a graph representation of the knotted structure using multi-cable Reidemeister and Node Deletion moves, as defined above. First, \algabbr{} performs a Reidemeister move to remove trivial crossings such as intra-cable loops and disambiguate the knot configuration. Next, \algabbr{} successively performs Node Deletion moves on the first non-trivial crossing with respect to the rightmost endpoint in the scene until no non-trivial crossings remain. \algabbr{} then performs Cable Extraction moves (discussed further in Section \ref{sec:method-manip}) to remove all cables from the scene and untangle any remaining trivial crossings. 

\subsection{Physical Disentangling}
\label{sec:hulk untangling}
\subsubsection{Perception}\label{subsec:perception}

\algabbr{} operates on a graph representation, which is not directly observable. Therefore, we instantiate \algabbr{} to operate on image inputs using learned perception components inspired by prior work~\cite{grannen2020untangling}. This makes it possible to instantiate \algabbr{} for cable disentangling given raw image input from a physical robotic system. As in HULK~\cite{grannen2020untangling}, we learn the newly-defined keypoints for holding and pulling actions and right and left endpoints from RGB image inputs to perform \algabbr{}'s multi-cable Reidemeister and Node Deletion moves on physical knots.

As in HULK, our approach uses a ResNet-34 backbone to learn two mappings, each of which transforms  an RGB image input to two heatmaps: 
(1) $g_1: \mathbb{R}^{640 \times 480 \times 3} \mapsto \mathbb{R}^{640 \times 480 \times 2}$ maps an image to 2 heatmaps centered respectively at the keypoints $\vc{\hat{p}}_\text{hold}$ and $\vc{\hat{p}}_\text{pull}$, located at the first non-trivial intersection as defined in \algabbr{}, and (2) $g_2: \mathbb{R}^{640 \times 480 \times 3} \mapsto \mathbb{R}^{640 \times 480 \times 2}$ maps an image to two heatmaps centered respectively at the 2 keypoints $\vc{\hat{p}}_r$ and $\vc{\hat{p}}_l$, respectively located at the endpoints of cables $c_i$ and $c_j$, $i \neq j$, to implement \algabbr{}'s multi-cable Reidemeister moves. Note that $\vc{\hat{p}}_r$ corresponds to the rightmost endpoint in the scene, while $\vc{\hat{p}}_l$ is the leftmost endpoint in the scene that does not belong to cable $c_i$. By querying LOKI as described in Section~\ref{sec:loki}, we obtain refined keypoints, $\vc{\tilde{p}}_l, \vc{\tilde{p}}_r, \vc{\tilde{p}}_{\text{hold}}, \vc{\tilde{p}}_{\text{pull}}$, and their corresponding gripper orientations for executing grasps.
Cable $c_i$ is \emph{semi-disentangled} from a cable configuration if the only crossings involving $c_i$ are trivial and $c_i$ can be fully disentangled when pulled apart from the other cables (see Fig. \ref{fig: labelling}).
We detect when the cable $c_i$ corresponding to the rightmost endpoint $\vc{\hat{p}}_r$ is semi-disentangled from the remaining cables with a binary classifier, $h: \mathbb{R}^{640 \times 480 \times 3} \mapsto \{0,1\}$.

\subsubsection{Manipulation}
\label{sec:method-manip}
To execute the actions from \algabbr{} on a set of physical cables, we apply a novel manipulation primitive, a \textbf{Cable Extraction move}, along with \textbf{Reidemeister moves} and \textbf{Node Deletion moves} (Sec.~\ref{sec:motion primitives}), to the task of multi-cable disentangling. The multi-cable Reidemeister and Node Deletion moves are executed as described in Section~\ref{sec:motion primitives} with the newly-defined keypoints described in Section~\ref{subsec:perception}. We iteratively disentangle each cable in the scene and drop fully disentangled cables at $\vc{p}_{\text{term}}$, a predefined point within the termination area.

Motivated by the slack management difficulties of multi-cable disentanglement (Sec.~\ref{sec:disentangling actions}), we define a \textbf{Cable Extraction move}, a novel manipulation primitive that fully disentangles and removes a semi-disentangled cable from a scene. We perform Cable Extraction moves when detecting that the cable $c_i$ corresponding to the rightmost endpoint $\vc{\tilde{p}}_r$ is semi-disentangled. The right arm grasps the semi-disentangled cable $c_i$ at its right endpoint $\vc{\tilde{p}}_r$, while the left arm \textit{pins} $\vc{\tilde{p}}_l$, the leftmost endpoint from cable $c_j$ ($i \neq j$), against the workspace surface without closing the gripper jaws and grasping the cable. Notably, we use \textit{soft pinning}, which differs from \textit{holding} in that a) in holding, the cable is not pushed against the workspace, and b) in soft pinning, the gripper jaws do not close to grasp the cable. Specifically, we use $\one_\mathrm{grasp} = 0$ to achieve soft pinning, which allows slack to slip through the jaws when only one cable is left in the scene and the predicted endpoints lie on the same cable. Then, the right arm pulls the cable $c_i$ to a predefined termination point $\vc{p}_{\text{term}}$, removing all trivial crossings:
\begin{flalign*}
\vc{a}_{t,r} &= (\tilde{p}_{x,r}, \tilde{p}_{y,r}, p_{x,\text{term}} -\tilde{p}_{x,r}, p_{y,\text{term}}-\tilde{p}_{y,r}, \hat{\theta}_r, 1) \\
\vc{a}_{t,l} &= (\tilde{p}_{x,l}, \tilde{p}_{y,l}, 0, 0, \hat{\theta}_l, 0).
\end{flalign*}

The full algorithm is summarized in Algorithm~\ref{alg:untangle_alg}. We start with a Reidemeister move to pull endpoints $\vc{\tilde{p}}_r$ and $\vc{\tilde{p}}_l$ to opposite ends of the workspace. Next, we successively perform Node Deletion moves with $\vc{\tilde{p}}_{\text{hold}}$ and $\vc{\tilde{p}}_{\text{pull}}$ at the first non-trivial undercrossing from the rightmost endpoint $\vc{\hat{p}}_r$. When we detect that $c_i$ is semi-disentangled, we perform Cable Extraction moves to undo any remaining trivial crossings involving $c_i$ and remove $c_i$ from the scene. This procedure continues until all cables are disentangled and deposited in the termination area. 

\begin{figure}[h!]
\MakeRobust{\Call}%
\begin{algorithm}[H]
\caption{Disentangling with \algabbr{}}
\label{alg:untangle_alg} 
\begin{algorithmic}[1]
\State \textbf{Input:} RGB image of cable
\State Predict $\vc{\tilde{p}}_l$, $\vc{\tilde{p}}_r$, $\vc{\tilde{p}}_{\text{hold}}$, $\vc{\tilde{p}}_{\text{pull}}$
\State $c_i, c_j \leftarrow$ cables corresponding to $\vc{\tilde{p}}_r, \vc{\tilde{p}}_l$, respectively.
\State Reidemeister move with $\vc{\tilde{p}}_r$ (cable $c_i$), $\vc{\tilde{p}}_l$ (cable $c_j$)
\While{workspace not empty}
\State Predict $\vc{\tilde{p}}_l$, $\vc{\tilde{p}}_r$, $\vc{\tilde{p}}_{\text{hold}}$, $\vc{\tilde{p}}_{\text{pull}}$
\State $c_i \leftarrow$ cable corresponding to $\vc{\tilde{p}}_r$
\State{Execute SPiDERMan recovery policy}
\If {cable $c_i$ is semi-disentangled} 
\State Cable Extraction move with $\vc{\tilde{p}}_r$, $\vc{\tilde{p}}_l$
\Else
\State Node Deletion move with $\vc{\tilde{p}}_{\text{hold}}$, $\vc{\tilde{p}}_{\text{pull}}$
\EndIf
\EndWhile
\State \textbf{return} DONE
\end{algorithmic}
\end{algorithm}
\end{figure}
\section{Experiments}\label{sec:experiments}
We evaluate \algabbr{} to disentangle knot configurations containing two or three cables and with three tiers of increasing difficulty. We implement the full system in experiments with the bilateral da Vinci surgical robot. Because this is the first work studying the multi-cable disentangling problem and because single-cable algorithms are not well-defined in this setting, we are not aware of any existing algorithms that would provide a meaningful baseline comparison.

\begin{table*}[!t]
\centering
\vspace{-0.1cm}
\begin{tabular}{|| c | c || c | c | c | c | c ||} 
\hline
Tier & Color & Success Rate & Disentangling Actions & Recovery Actions & Total Actions & Failure Modes \\ 
\hline
\hline
1 & r-w & 12/12 & 7 & 0 & 7.5 & A (0), B (0), C (0), D (0) \\
\hline
1 & w-w & 10/12 & 11.5 & 1 & 12.5 & A (0), B (0), C (1), D (1) \\
\hline
2 & r-w & 7/12 & 19 & 0 & 20 & A (1), B (1), C (1), D (2) \\
\hline
2 & w-w & 9/12 & 15.5 & 2 & 15 & A (0), B (1), C (2), D (0) \\
\hline
3 & r-w-w & 11/12 & 16 & 1 & 17 & A (0), B (0), C (0), D (1) \\
\hline
3 & w-w-w & 9/12 & 15 & 1 & 16 & A (0), B (1), C (0), D (2) \\
\hline
\end{tabular}
\\
\caption{\textbf{Physical Experiment Results:} We report the success rate and median number of actions to fully disentangle all cables in a scene using the physical implementation of \algabbr{}. We consider sets of cables that are all similarly colored (white-white) and differently colored (red-white). The \algabbr{} implementation disentangles two cables in Tier 1 and 2 configurations and three cables in Tier 3 configurations with an overall success rate of 80.5\%. We observe four failure modes: (A) one or more cables springing out of the manipulation workspace, (B) gripper collision in high-density configurations, (C) exceeding the 20-30 maximum number of disentangling actions, and (D) moving entangled cables to the termination area.}
\label{table:phys_exp_results}
\vspace{-0.3cm}
\end{table*}

\subsection{Training Dataset Generation}
We train the Reidemeister and Node Deletion coarse keypoint prediction models $g_1$ and $g_2$ on a dataset of 270 real workspace images with hand-labeled keypoints, augmented to a dataset of 7,020 examples via affine transforms, lighting shifts, and blurring. We similarly train the semi-disentanglement classifier $h$ on a dataset of 170 real workspace images. For each image, we assign labels 0 or 1 by hand to indicate    ``semi-disentangled" or ``entangled," respectively, augment to a dataset of 5,200 examples using similar augmentation techniques as mentioned above, and train with a binary cross-entropy loss. All datasets consist \emph{only} of configurations with up to two cables, where the cables' colors are either both white or red and white. To reduce manipulation errors during experiments, we project all keypoint predictions $\vc{\hat{p}} \in \{\vc{\hat{p}}_\text{hold}, \vc{\hat{p}}_\text{pull}, \vc{\hat{p}}_l, \vc{\hat{p}}_r\}$ onto the cable mask obtained by color thresholding from the background.

We train LOKI on 3,000 $200 \times 200$ crops of images of configurations with one red and one white cable, generated in simulation via Blender 2.80~\cite{lallemand1998blender}. Offset heatmap labels are generated in simulation as 2D Gaussian distributions centered along the cable width with a standard deviation of 5 px. SPiDERMan detects when to perform recovery actions via analytical methods for sensing cable contours~\cite{nonplanar-cable-untangling}. 

\subsection{Tiers of Difficulty}
\begin{figure} 
\centering
\includegraphics[width=\linewidth]{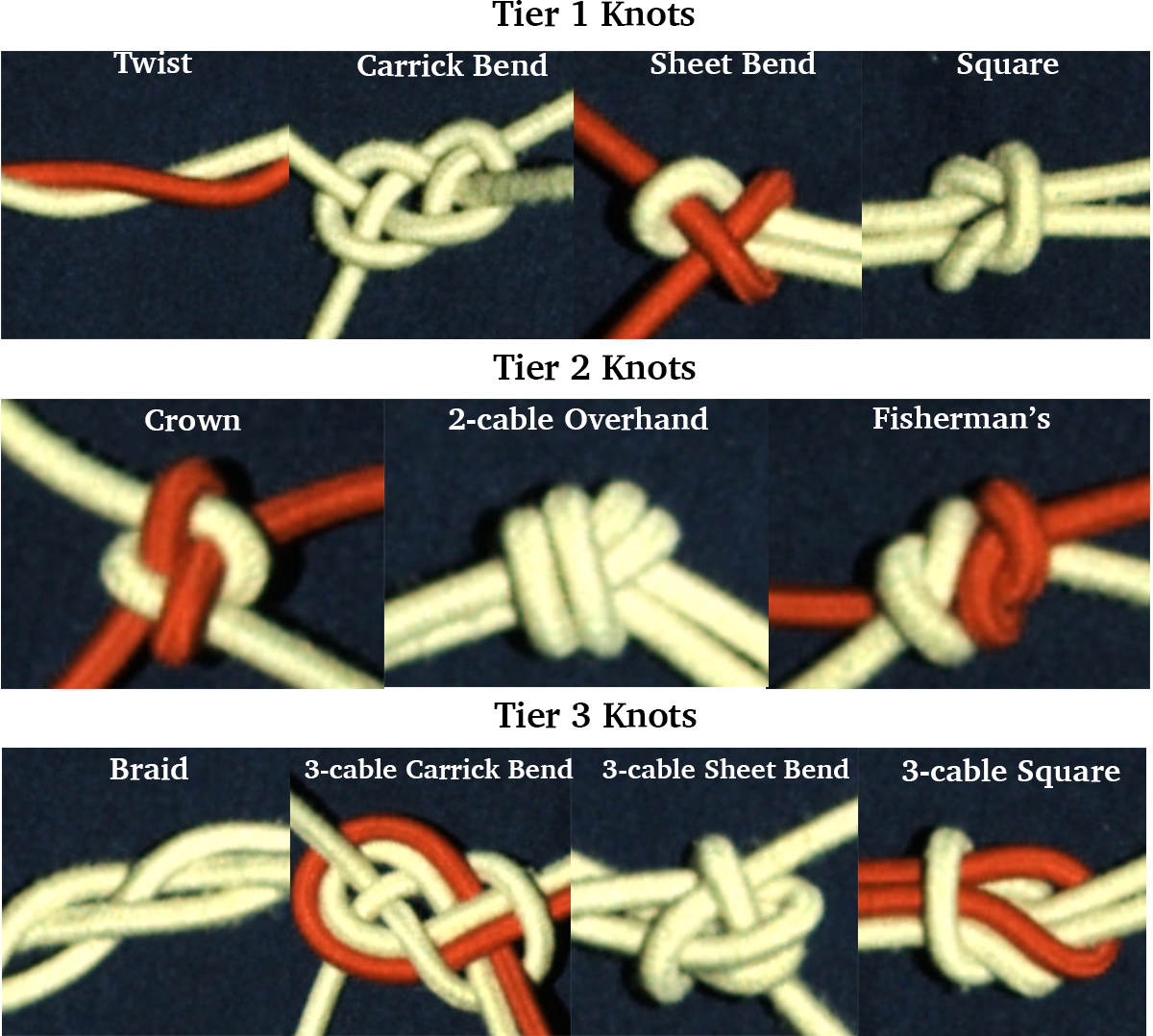}
\caption{\textbf{Tiers of Configuration Difficulty}: The configurations considered in this paper are organized into 3 tiers. Tiers 2 and 3 include novel knots not in the training set. For each configuration, we consider both settings where cables have contrasting colors (red and white) and where cables are of the same color (all white).}
\label{fig:all_knots}
\end{figure}

Across all difficulty tiers, the knot types considered are dense (Figure~\ref{fig:all_knots}). The tiers are defined by the number of cables in the knot and whether the class of knots was present in the training dataset. In all three tiers, we consider knots in which cables have contrasting colors (red and white) and knots in which all cables are the same color (white).

\textbf{Tier 1: } Two-cable knots where the class of knots was present in the training dataset (two-cable twists, Carrick Bend, Sheet Bend, and Square knots).

\textbf{Tier 2: } Two-cable knots where the class of knots was not present in the training dataset (Crown, Fisherman's, and two-cable Overhand knots). 

\textbf{Tier 3: } Three-cable knots where the class of knots was not present in the training dataset (braids, three-cable Carrick Bend, three-cable Sheet Bend, and three-cable Square knots).

\subsection{Experimental Setup}
We use a da Vinci Research Kit (dVRK) surgical robot with two 7-DOF arms and 2 or 3 cut elastic hairties of diameter 5 mm and length 15 cm. Hairties have a smooth surface and fit well in the dVRK's end effector. We also use a foam padded stage on which the hairties rest during experiments to avoid end effector damage under collisions with the manipulation surface and to create friction with the hairties to prevent them from sliding out of the workspace. 
We collect $1920 \times 1200$ overhead RGB images for perception inference with a Zivid OnePlus RGBD camera. 

\subsection{Results}\label{subsec:results}
Table~\ref{table:phys_exp_results} presents the results from the physical trials. \algabbr{} succeeds in disentangling all cables with success rates of $91.6\%$, $66.6\%$, and $83.3\%$ on Tier 1, Tier 2, and Tier 3 configurations, respectively. The success rates drop for the knots in Tiers 2 and 3, as expected since they are not present in training data images. In successful Tier 2 and 3 cases, both the numbers of disentangling actions (Node Deletion and Cable Extraction moves) and SPiDERMan Recovery actions (Re-posing and Wedge Recovery moves) increase compared to Tier 1.

\subsection{Failure Modes}\label{subsec:failure_modes}
In the disentangling experiments, the physical implementation of \algabbr{} encounters four failure modes:
\begin{enumerate}[(A)]
    \item One or more cables springing out of the reachable manipulation workspace due to elastic cable physics.
    \item Robot gripper jaws colliding when executing Node Deletion moves in high-density cable configurations.
    \item Exceeding a maximum threshold of 20, 30, and 30 disentangling actions for Tiers 1, 2, and 3 respectively, due to repeatedly-poor action predictions and executions. We allow fewer total actions in Tier 1 because the class of Tier 1 knots was seen in training.
    \item Semi-disentangled cables dropped in the termination area---rather than fully disentangled cables---due to poorly-executed Cable Extraction moves that do not effectively pin down remaining cables in the scene.
\end{enumerate}
We observe that the most common failure mode is (D). When the left arm does not effectively pin the remaining cables in the scene, high cable friction causes multiple cables to move with the grasped semi-disentangled cable into the termination area without removing trivial crossings. 
We also observe two manipulation failure modes relating to cable grasps. Due to the hairties' elastic properties, poor cable grasps occasionally cause one or more cables to spring out of the reachable manipulation workspace, which yields an irrecoverable state (A). In high-density configurations, disentangling actions often require gripper jaws to grasp adjacent cable segments that are very close together. These grasps may cause gripper jaw collision, which requires human intervention to reset the robot and is deemed a failure (B).
We limit the number of disentangling actions to 20 actions for Tier 1 configurations and 30 actions for Tier 2 and 3 configurations to end rollouts when the robot is repeatedly unable to execute effective disentangling actions in pathological high-density configurations where gripper jaws cannot grasp between cable segments (C).  

\section{Discussion and Future Work}
This work formalizes the problem of autonomously disentangling multiple cables and presents \algabbr{}, a geometric algorithm for disentangling multi-cable knots. \algabbr{} iteratively undoes inter-cable and intra-cable crossings by operating on a graphical knot representation. We build on perception-driven approaches from prior work (HULK, LOKI, SPiDERMan) to instantiate \algabbr{} for disentangling cables on a physical robotic system using image inputs. Experiments suggest that the physical implementation of \algabbr{} can disentangle up to three cables with 80.5\% success.

In future work, we will explore how \algabbr{} and its physical implementation can be extended to differing cable sizes and textures. We will also explore \algabbr{} in scenarios where one endpoint is fixed or where cables are tangled with rigid objects such as electrical appliances or tools.

\section*{Acknowledgements}
\small
This research was performed at the AUTOLAB at UC Berkeley in affiliation with the Berkeley AI Research (BAIR) Lab, the CITRIS “People and Robots” (CPAR) Initiative, and the Real-Time Intelligent Secure Execution (RISE) Lab. The authors were supported in part by donations from Toyota Research Institute and by equipment grants from PhotoNeo, NVidia, and Intuitive Surgical.


\printbibliography

@STRING{isrr = {Int. S. Robotics Research (ISRR)}}

@inproceedings{sundaresan2020learning,
  title={Learning rope manipulation policies using dense object descriptors trained on synthetic depth data},
  author={Sundaresan, Priya and Grannen, Jennifer and Thananjeyan, Brijen and Balakrishna, Ashwin and Laskey, Michael and Stone, Kevin and Gonzalez, Joseph E and Goldberg, Ken},
  booktitle={2020 IEEE International Conference on Robotics and Automation},
  pages={9411--9418},
  year={2020},
  organization={IEEE}
}

@inproceedings{nair2017combining,
  title={Combining self-supervised learning and imitation for vision-based rope manipulation},
  author={Nair, Ashvin and Chen, Dian and Agrawal, Pulkit and Isola, Phillip and Abbeel, Pieter and Malik, Jitendra and Levine, Sergey},
  booktitle={2017 IEEE International Conference on Robotics and Automation},
  pages={2146--2153},
  year={2017},
  organization={IEEE}
}

@article{yan2020self,
  title={Self-Supervised Learning of State Estimation for Manipulating Deformable Linear Objects},
  author={Yan, Mengyuan and Zhu, Yilin and Jin, Ning and Bohg, Jeannette},
  journal={IEEE Robotics and Automation Letters},
  volume={5},
  number={2},
  pages={2372--2379},
  year={2020},
  publisher={IEEE}
}

@inproceedings{pathak2018zero,
  title={Zero-shot visual imitation},
  author={Pathak, Deepak and Mahmoudieh, Parsa and Luo, Guanghao and Agrawal, Pulkit and Chen, Dian and Shentu, Yide and Shelhamer, Evan and Malik, Jitendra and Efros, Alexei A and Darrell, Trevor},
  booktitle={Proceedings of the IEEE conference on computer vision and pattern recognition workshops},
  pages={2050--2053},
  year={2018}
}

@article{lin2020softgym,
  title={{SoftGym}: Benchmarking Deep Reinforcement Learning for Deformable Object Manipulation},
  author={Lin, Xingyu and Wang, Yufei and Olkin, Jake and Held, David},
  journal={Conference on Robot Learning},
  year={2020}
}

@article{hoquevisuospatial,
  title={VisuoSpatial Foresight for Multi-Step, Multi-Task Fabric Manipulation},
  author={Hoque, Ryan and Seita, Daniel and Balakrishna, Ashwin and Ganapathi, Aditya and Tanwani, Ajay Kumar and Jamali, Nawid and Yamane, Katsu and Iba, Soshi and Goldberg, Ken}
}

@article{ebert2018visual,
  title={Visual foresight: Model-based deep reinforcement learning for vision-based robotic control},
  author={Ebert, Frederik and Finn, Chelsea and Dasari, Sudeep and Xie, Annie and Lee, Alex and Levine, Sergey},
  journal={arXiv preprint arXiv:1812.00568},
  year={2018}
}

@article{zhang2020robots,
  title={Robots of the Lost Arc: Learning to Dynamically Manipulate Fixed-Endpoint Ropes and Cables},
  author={Zhang, Harry and Ichnowski, Jeffrey and Seita, Daniel and Wang, Jonathan and Goldberg, Ken},
  journal={IEEE International Conference on Robotics and Automation},
  year={2021}
}

@article{billard2019trends,
  title={Trends and challenges in robot manipulation},
  author={Billard, Aude and Kragic, Danica},
  journal={Science},
  volume={364},
  number={6446},
  year={2019},
  publisher={American Association for the Advancement of Science}
}

@inproceedings{herguedas2019survey,
  title={Survey on multi-robot manipulation of deformable objects},
  author={Herguedas, Rafael and L{\'o}pez-Nicol{\'a}s, Gonzalo and Arag{\"u}{\'e}s, Rosario and Sag{\"u}{\'e}s, Carlos},
  booktitle={2019 24th IEEE International Conference on Emerging Technologies and Factory Automation},
  year={2019},
  organization={IEEE}
}

@inproceedings{chi2019occlusion,
  title={Occlusion-robust Deformable Object Tracking without Physics Simulation},
  author={Chi, Cheng and Berenson, Dmitry},
  booktitle={2019 IEEE/RSJ International Conference on Intelligent Robots and Systems},
  pages={6443--6450},
  year={2019},
  organization={IEEE}
}

@inproceedings{lui2013tangled,
  title={Tangled: Learning to untangle ropes with {RGB-D} perception},
  author={Lui, Wen Hao and Saxena, Ashutosh},
  booktitle={2013 IEEE/RSJ International Conference on Intelligent Robots and Systems},
  pages={837--844},
  year={2013},
  organization={IEEE}
}

@article{grannen2020untangling,
  title={Untangling Dense Knots by Learning Task-Relevant Keypoints},
  author={Grannen, Jennifer and Sundaresan, Priya and Thananjeyan, Brijen and Ichnowski, Jeffrey and Balakrishna, Ashwin and Hwang, Minho and Viswanath, Vainavi and Laskey, Michael and Gonzalez, Joseph E and Goldberg, Ken},
  journal={Conference on Robot Learning},
  year={2020}
}

@article{lee2020learning,
  title={Learning Arbitrary-Goal Fabric Folding with One Hour of Real Robot Experience},
  author={Lee, Robert and Ward, Daniel and Cosgun, Akansel and Dasagi, Vibhavari and Corke, Peter and Leitner, Jurgen},
  journal={Conference on Robot Learning},
  year={2020}
}

@article{seita2020learning,
  title={Learning to Rearrange Deformable Cables, Fabrics, and Bags with Goal-Conditioned Transporter Networks},
  author={Seita, Daniel and Florence, Pete and Tompson, Jonathan and Coumans, Erwin and Sindhwani, Vikas and Goldberg, Ken and Zeng, Andy},
  journal={IEEE International Conference on Robotics and Automation},
  year={2021}
}

@inproceedings{florence2018dense,
  title={Dense Object Nets: Learning Dense Visual Object Descriptors By and For Robotic Manipulation},
  author={Florence, Peter R and Manuelli, Lucas and Tedrake, Russ},
  booktitle={Conference on Robot Learning},
  pages={373--385},
  year={2018},
  organization={PMLR}
}

@article{ganapathi2020learning,
  title={Learning Dense Visual Correspondences in Simulation to Smooth and Fold Real Fabrics},
  author={Ganapathi, Aditya and Sundaresan, Priya and Thananjeyan, Brijen and Balakrishna, Ashwin and Seita, Daniel and Grannen, Jennifer and Hwang, Minho and Hoque, Ryan and Gonzalez, Joseph E and Jamali, Nawid and others},
  journal={IEEE International Conference on Robotics and Automation},
  year={2021}
}

@article{seita2019deep,
  title={Deep imitation learning of sequential fabric smoothing policies},
  author={Seita, Daniel and Ganapathi, Aditya and Hoque, Ryan and Hwang, Minho and Cen, Edward and Tanwani, Ajay Kumar and Balakrishna, Ashwin and Thananjeyan, Brijen and Ichnowski, Jeffrey and Jamali, Nawid and others},
  journal={2020 IEEE/RSJ International Conference on Intelligent Robots and Systems},
  year={2020}
}

@article{mayer2008system,
  title={A system for robotic heart surgery that learns to tie knots using recurrent neural networks},
  author={Mayer, Hermann and Gomez, Faustino and Wierstra, Daan and Nagy, Istvan and Knoll, Alois and Schmidhuber, J{\"u}rgen},
  journal={Advanced Robotics},
  volume={22},
  number={13-14},
  pages={1521--1537},
  year={2008},
  publisher={Taylor \& Francis}
}

@inproceedings{van2010superhuman,
  title={Superhuman performance of surgical tasks by robots using iterative learning from human-guided demonstrations},
  author={Van Den Berg, Jur and Miller, Stephen and Duckworth, Daniel and Hu, Humphrey and Wan, Andrew and Fu, Xiao-Yu and Goldberg, Ken and Abbeel, Pieter},
  booktitle={2010 IEEE International Conference on Robotics and Automation},
  year={2010},
  organization={IEEE}
}

@inproceedings{yamakawa2007one,
  title={One-handed knotting of a flexible rope with a high-speed multifingered hand having tactile sensors},
  author={Yamakawa, Yuji and Namiki, Akio and Ishikawa, Masatoshi and Shimojo, Makoto},
  booktitle={2007 IEEE/RSJ International Conference on Intelligent Robots and Systems},
  pages={703--708},
  year={2007},
  organization={IEEE}
}

@article{sanchez2018robotic,
  title={Robotic manipulation and sensing of deformable objects in domestic and industrial applications: a survey},
  author={Sanchez, Jose and Corrales, Juan-Antonio and Bouzgarrou, Belhassen-Chedli and Mezouar, Youcef},
  journal={The International Journal of Robotics Research},
  volume={37},
  number={7},
  pages={688--716},
  year={2018},
  publisher={SAGE Publications Sage UK: London, England}
}

@article{seita2018deep,
  title={Deep transfer learning of pick points on fabric for robot bed-making},
  author={Seita, Daniel and Jamali, Nawid and Laskey, Michael and Tanwani, Ajay Kumar and Berenstein, Ron and Baskaran, Prakash and Iba, Soshi and Canny, John and Goldberg, Ken},
  journal={International Symposium on Robotics Research (ISRR)},
  year={2019}
}

@inproceedings{merlet2010portable,
  title={A portable, modular parallel wire crane for rescue operations},
  author={Merlet, Jean-Pierre and Daney, David},
  booktitle={2010 IEEE International Conference on Robotics and Automation},
  year={2010},
  organization={IEEE}
}

@misc{kempf2011fiber,
  title={Fiber cable made of high-strength synthetic fibers for a helicopter rescue winch},
  author={Kempf, Florian and Fischer, Juergen},
  year={2011},
  month=jan,
  publisher={Google Patents},
  note={US Patent 7,866,245}
}

@article{nonplanar-cable-untangling,
  title={Untangling Dense Non-Planar Knots by Learning Manipulation Features and Recovery Policies},
  author={Sundaresan, Priya and Grannen, Jennifer and Thananjeyan, Brijen and Balakrishna, Ashwin and Ichnowski, Jeff and Novoseller, Ellen and Hwang, Minho and Laskey, Michael and Gonzalez, Joseph E. and Goldberg, Ken},
  journal={Robotics: Science and Systems},
  year={2021}
}

@misc{lallemand1998blender,
  title={Blender},
  author={Lallemand, Thomas},
  year={1998},
  month=jun,
  publisher={Google Patents},
  note={US Patent App. 29/074,220}
}

\clearpage



\end{document}